\documentclass[sigconf]{acmart}
\usepackage{url}
\usepackage{graphicx}
\usepackage{amsmath}
\usepackage{booktabs}
\usepackage{xcolor}
\usepackage{multirow} 
\usepackage{diagbox}
\usepackage{enumitem}
\usepackage{subcaption}
\usepackage{anyfontsize}
\usepackage{float}
\usepackage{etoolbox}
\usepackage{lipsum}

\usepackage{booktabs}
\usepackage{tabularx}
\newcolumntype{Y}{>{\raggedright\arraybackslash}X}

\makeatletter
\DeclareRobustCommand\onedot{\futurelet\@let@token\@onedot}
\def\@onedot{\ifx\@let@token.\else.\null\fi\xspace}

\usepackage[normalem]{ulem}
\useunder{\uline}{\ul}{}

\AtBeginDocument{%
  }

\copyrightyear{2025}
\acmYear{2025}
\setcopyright{acmlicensed}\acmConference[MM '25]{Proceedings of the 33rd ACM International Conference on Multimedia}{October 27--31, 2025}{Dublin, Ireland}
\acmBooktitle{Proceedings of the 33rd ACM International Conference on Multimedia (MM '25), October 27--31, 2025, Dublin, Ireland}
\acmDOI{10.1145/3746027.3762067}
\acmISBN{979-8-4007-2035-2/2025/10}

\begin{document}

\title{Event-Enriched Image Analysis Grand Challenge at ACM Multimedia 2025}


\author{Thien-Phuc Tran}
\authornote{Both authors contributed equally to this research.}
\orcid{0009-0008-0800-6884}
\affiliation{%
  \institution{University of Science,}
  \city{VNU-HCM}
  \country{Vietnam}
}
\author{Minh-Quang Nguyen}
\authornotemark[1]
\orcid{0009-0008-3520-4624}
\affiliation{%
  \institution{University of Science,}
  \city{VNU-HCM}
  \country{Vietnam}
}
\author{Minh-Triet Tran}
\orcid{0000-0003-3046-3041}
\affiliation{%
  \institution{University of Science,}
  \city{VNU-HCM}
  \country{Vietnam}
}
\author{Tam V. Nguyen}
\orcid{0000-0003-0236-7992}
\affiliation{%
  \institution{University of Dayton,}
  \state{Ohio}
  \country{United States}
}
\author{Trong-Le Do}
\orcid{0000-0002-2906-0360}
\affiliation{%
  \institution{University of Science,}
  \city{VNU-HCM}
  \country{Vietnam}
}
\author{Duy-Nam Ly}
\orcid{0000-0003-4304-2334}
\affiliation{%
  \institution{University of Science,}
  \city{VNU-HCM}
  \country{Vietnam}
}
\author{Viet-Tham Huynh}
\orcid{0000-0002-8537-1331}
\affiliation{%
  \institution{University of Science,}
  \city{VNU-HCM}
  \country{Vietnam}
}
\author{Khanh-Duy Le}
\orcid{0000-0002-8297-5666}
\affiliation{%
  \institution{University of Science,}
  \city{VNU-HCM}
  \country{Vietnam}
}
\author{Mai-Khiem Tran}
\orcid{0000-0001-5460-0229}
\affiliation{%
  \institution{University of Science,}
  \city{VNU-HCM}
  \country{Vietnam}
}
\author{Trung-Nghia Le}
\orcid{0000-0002-7363-2610}
\affiliation{%
  \institution{University of Science,}
  \city{VNU-HCM}
  \country{Vietnam}
}
\authornote{Corresponding author. Email: ltnghia@fit.hcmus.edu.vn}


\renewcommand{\shortauthors}{T.-P. Tran et al.}


\begin{abstract}

The Event-Enriched Image Analysis (EVENTA) Grand Challenge, hosted at ACM Multimedia 2025, introduces the first large-scale benchmark for event-level multimodal understanding. Traditional captioning and retrieval tasks largely focus on surface-level recognition of people, objects, and scenes, often overlooking the contextual and semantic dimensions that define real-world events. EVENTA addresses this gap by integrating contextual, temporal, and semantic information to capture the who, when, where, what, and why behind an image. Built upon the OpenEvents V1 dataset, the challenge features two tracks: Event-Enriched Image Retrieval and Captioning, and Event-Based Image Retrieval. A total of 45 teams from six countries participated, with evaluation conducted through Public and Private Test phases to ensure fairness and reproducibility. The top three teams were invited to present their solutions at ACM Multimedia 2025. EVENTA establishes a foundation for context-aware, narrative-driven multimedia AI, with applications in journalism, media analysis, cultural archiving, and accessibility. Further details about the challenge are available at the official homepage: \url{https://ltnghia.github.io/eventa/eventa-2025}.

\end{abstract}




\maketitle


\section{Introduction}

Recent advances in multimodal AI have significantly improved image captioning~\cite{xu2015show}, object detection~\cite{girshick2014rich}, and semantic segmentation~\cite{long2015fully}, enabling systems to describe visible content with high accuracy. Nevertheless, these methods largely remain limited to surface-level recognition of people, objects, and actions~\cite{biten2019good}. In real-world applications, such literal descriptions are insufficient. For example, identifying that ``people are watching television'' leaves critical contextual questions unanswered: what program are they watching, is it culturally or politically significant, and why are they enjoying it? Current AI systems rarely address such deeper dimensions of meaning, resulting in captions that lack explanatory depth and narrative coherence~\cite{krause2017hierarchical}.

The Event-Enriched Image Analysis (EVENTA) Grand Challenge addresses this gap by shifting the focus from visual analysis to event-level understanding. Its central objective is to transform image interpretation into contextualized storytelling that explains not only what is seen but also who is involved, when and where the event takes place, what is happening, and why it is significant. Achieving this requires combining visual information with contextual reasoning, thereby producing narrative-rich captions that situate images within broader social, historical, or cultural frames, thus providing the audience with deeper insights into the semantics underlying the perceived images and textual content. 

By introducing this event-aware perspective, our EVENTA challenge advances a model of perception that captures not only visual elements but also their contextual significance. Understanding an image becomes an exercise in reconstructing its context, implications, and human relevance, rather than merely enumerating objects and scenes. For instance, a protest in a city square, a historic moment captured in a news photograph, or a family gathering suffused with subtle emotions each demands interpretations that extend beyond literal recognition. Accordingly, our challenge highlights the need for richer semantic descriptions that capture not only the visible signals but also their deeper event-centric meaning. The implications of this paradigm extend across multiple domains. In journalism and media analysis, event-enriched captioning can support automatic generation of accurate and informative descriptions for archival and contemporary news imagery. For event discovery and retrieval, the integration of contextual knowledge enables more expressive and semantically meaningful search capabilities. In cultural archiving and storytelling, such methods provide tools for preserving socio-historical moments with narrative fidelity.


To this end, the EVENTA 2025 Grand Challenge promotes the development of context-aware captions that incorporate event-specific details such as names, timelines, and outcomes. By providing a benchmark that evaluates semantic depth, our challenge offers the community a testbed for advancing multimodal systems beyond descriptive accuracy toward genuine contextual understanding. The EVENTA challenge attracted broad international participation, with 45 teams from six countries registering across its two tracks: Event-Enriched Image Retrieval and Captioning and Event-Based Image Retrieval. The evaluation was conducted in two stages: a Public Test phase for system development and tuning, followed by a Private Test phase to ensure fair and unbiased assessment. Based on the final leaderboard, the top three teams were invited to present their solutions at ACM Multimedia 2025, recognizing their outstanding contributions to advancing event-aware multimodal understanding. 

The success of EVENTA 2025 demonstrates not only the feasibility of integrating vision, retrieval, and contextual reasoning but also the growing interest of the community in moving beyond surface-level recognition. By presenting a benchmark for event-enriched captioning and retrieval, EVENTA offers a basis for subsequent studies that connect computer vision, natural language processing, and information retrieval. The results indicate the potential of context-rich, interpretive approaches to visual media and open avenues for developing AI systems oriented toward understanding not just what is shown, but also who is involved, when and where it occurs, and why it matters.


\section{Related Work}

\subsection{Image Captioning}
Recent advances in image captioning leverage large-scale vision--language pretraining and unified architectures. Contrastive models like CLIP demonstrated the power of aligning visual and textual embeddings on web-scale data, enabling strong zero-shot recognition and retrieval \citep{pmlr-v139-radford21a}. Building on this, transformer-based frameworks (e.g., ALBEF) first align image--text representations and then fuse them with cross-modal attention; its align-before-fuse strategy with momentum distillation achieved state-of-the-art results on captioning and multimodal understanding \citep{li2021albef}. BLIP unified vision--language understanding and generation by bootstrapping captions for noisy web images through a captioner-and-filter scheme and a flexible encoder--decoder, improving COCO retrieval and captioning benchmarks \citep{pmlr-v162-li22n}. Generative transformer models have also emerged: GIT simplifies the architecture to one ViT encoder and one text decoder under a language-modeling objective, scaling pretraining to surpass human-level CIDEr on TextCaps \citep{wang2022git}.

Another line of work focuses on multimodal reasoning and external knowledge to produce richer, context-aware captions. Flamingo bridges pretrained visual encoders and large language models via novel cross-attention mechanisms, enabling in-context learning for open-ended vision--language tasks without task-specific fine-tuning \citep{alayrac2022flamingo}. Retrieval-augmented captioning injects real-world context: EXTRA conditions a captioner on top-$k$ captions retrieved from a datastore and encodes image + retrieved text with a V\&L encoder, significantly improving quality on COCO \citep{ramos2023extra}. In the news domain, Visually-Aware Context Modeling integrates article signals (e.g., face--name cues) to guide caption generation, enabling naming and event details beyond the visible content \citep{qu2024news}. EVCap augments LLM-based captioning with an external visual--name memory to improve open-world comprehension and novel-object grounding \citep{li2024evcap}. Despite progress, models often miss event-level semantics (specific names, temporal context, implications) not directly visible, motivating event-grounded captioning that goes beyond surface descriptions.

\subsection{Image Retrieval}
Image--text retrieval has similarly advanced via powerful vision--language representations and large datasets. Dual-encoder models trained on millions to billions of pairs (e.g., CLIP, ALIGN) learn a shared embedding space that enables effective cross-modal retrieval, even zero-shot \citep{pmlr-v139-radford21a,pmlr-v139-jia21b}. Subsequent methods improved fine-grained alignment: ALBEF’s contrastive align-before-fuse pretraining yields strong retrieval with modest data \citep{li2021albef}, and BLIP’s caption filtering further boosts COCO recall \citep{pmlr-v162-li22n}. More recently, ALIP integrates synthetic captions to mitigate web noise, achieving state-of-the-art zero-shot image--text retrieval \citep{yang2023alip}.

Beyond generic short-query retrieval, event-centric queries require aligning narrative descriptions with visual scenes. Systems increasingly consider contextual cues (articles, names, timelines) and leverage language models to parse complex queries; yet long, narrative queries remain challenging, highlighting the need for event-grounded retrieval capable of bridging vision and storytelling---precisely the focus of EVENTA-style tasks \citep{qu2024news,ramos2023extra,li2024evcap}.

Motivated by the gaps in current multimodal understanding research, we organize the EVENTA Grand Challenge at ACM Multimedia 2025. The tasks are designed to encourage participants to explore deeper layers of meaning beyond visible content, underscoring the need for event-centric understanding of multimedia and fostering the development of systems for event-based downstream tasks.

\section{Challenge Description}

\subsection{Tasks}

\subsubsection{Track 1: Event-Enriched Image Retrieval and Captioning}

\begin{figure}[t!]
    \centering
    \includegraphics[width=\linewidth]{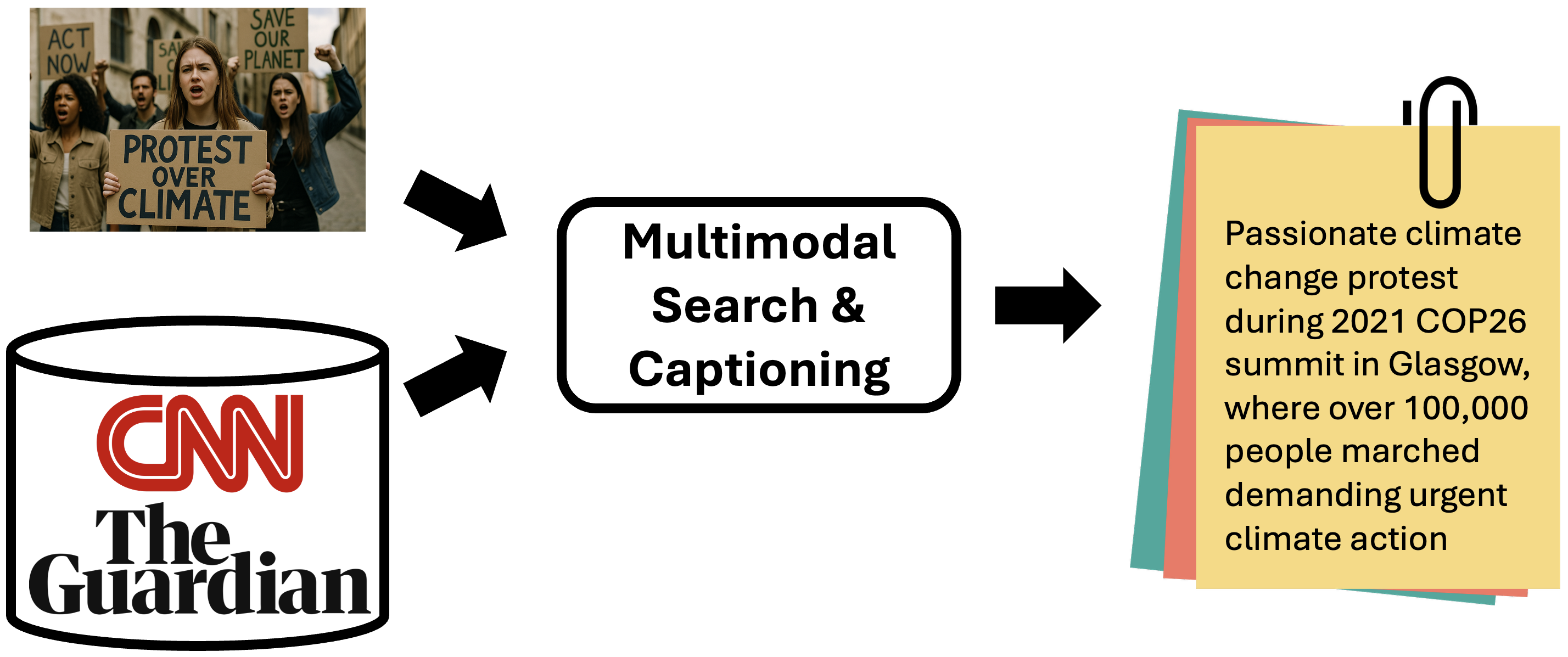}
    \vspace{-5mm}
    \caption{Given a dataset of news, including images and content, with a set of query images. The task is to utilize multimodal search for retrieving related figures and apply captioning models to generate descriptions for each query.}
    \Description[eventa track 1]{eventa track 1}
    \label{fig:eventa25-track-1}
\end{figure}

This track is designed to generate captions that deliver richer and more comprehensive accounts of an image (\autoref{fig:eventa25-track-1}). Unlike conventional captioning tasks that focus solely on surface-level visual elements, the objective here is to produce context-aware descriptions that incorporate entity names, attributes, temporal and spatial information, event outcomes, and other details that cannot be inferred from visual content alone. Given an input image, participants are required to retrieve relevant evidence from a curated external article database and integrate this information into the caption generation process. By adopting a retrieval-augmented generation (RAG) paradigm, this track promotes the creation of coherent, narrative-driven captions that capture not only the visible elements of a scene but also its broader context and significance. The resulting captions therefore move beyond description toward event-level understanding, offering a more complete and semantically grounded interpretation of what the image represents.

\begin{figure}[t!]
    \centering
    \includegraphics[width=\linewidth]{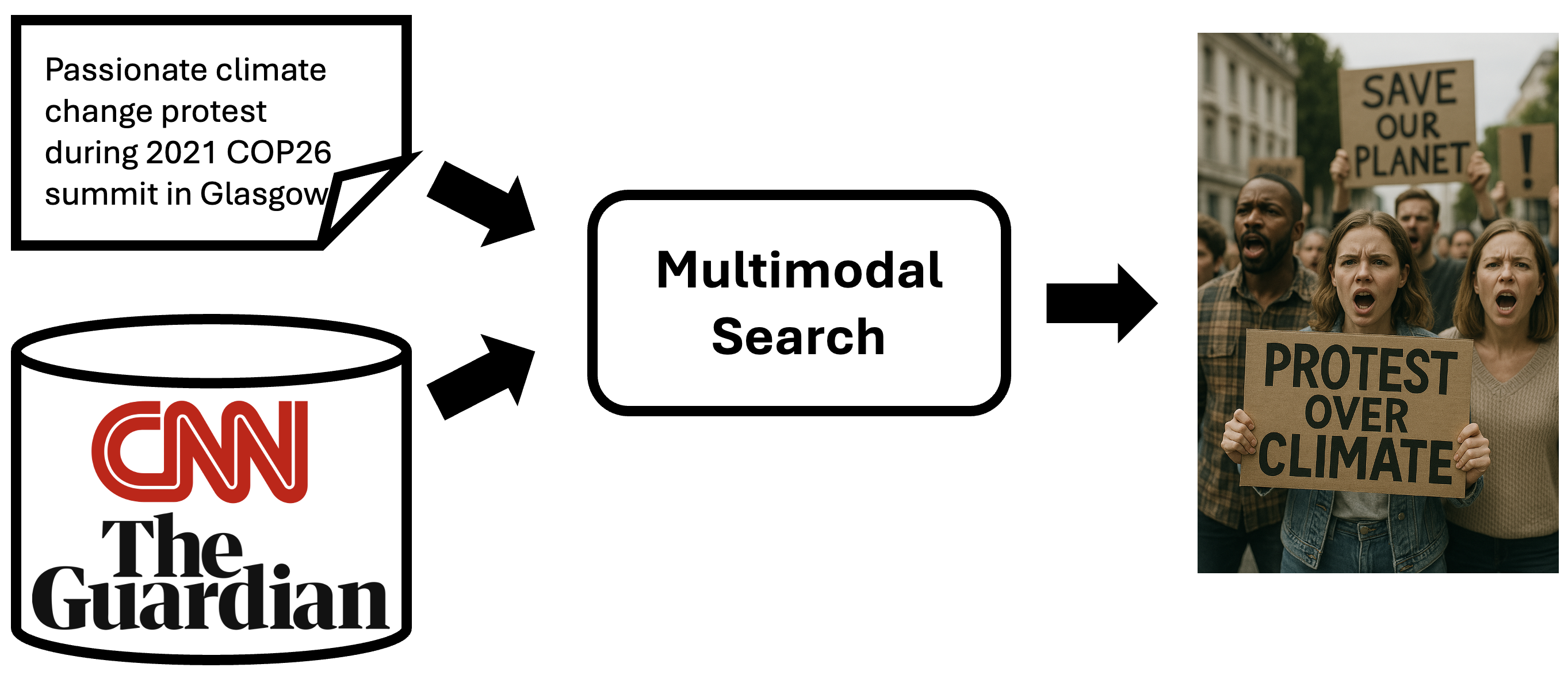}
    \vspace{-5mm}
    \caption{This task involves mainly image retrieval by browsing the most related image from the dataset for each query description, given a dataset of news, including images and content.}
    \label{fig:eventa25-track-2}
    \Description[eventa track 2]{eventa track 2}
\end{figure}

\subsubsection{Track 2: Event-Based Image Retrieval}
In this track, participants are provided with a realistic event-related caption and are required to retrieve the corresponding images from a curated database (\autoref{fig:eventa25-track-2}). This task represents a fundamental problem at the intersection of computer vision and natural language processing, requiring the construction of a joint representation space in which visual and textual modalities can be effectively aligned and compared. Text-based image retrieval has broad applications, including search engines, medical imaging, e-commerce, and digital asset management. Despite substantial progress, several challenges remain, such as addressing abstract or ambiguous queries, scaling retrieval systems to large datasets, and ensuring robustness to linguistic variability and biases in training data. By situating retrieval within the context of real-world event information, this track emphasizes the need for models that can move beyond surface-level matching toward event-aware retrieval, thereby advancing both the accuracy and contextual relevance of cross-modal search.

\begin{table}[t!]
  \centering
  \small
  \caption{Evaluation metrics used in Track 1: Event-Enriched Image Retrieval and Image Captioning.}
  \vspace{-3mm}
  \label{tab:metrics_task1}
  \begin{tabularx}{\columnwidth}{@{} l Y l @{}}
    \toprule
    \textbf{Metrics}  & \textbf{Description}                                            & \textbf{Domain}   \\
    \midrule
    AP              & Average Precision—measures retrieval precision across thresholds  & Retrieval         \\
    Recall@1        & Whether the correct item is ranked at the top                     & Retrieval         \\
    Recall@10       & Whether the correct item is within the top 10 results             & Retrieval         \\
    CLIPScore ~\cite{hessel2021clipscore}      & Semantic alignment between image and caption                      & Captioning        \\
    CIDEr  ~\cite{Vedantam_2015_CVPR}         & Agreement between generated and reference captions                & Captioning        \\
    \bottomrule
  \end{tabularx}
\end{table}

\begin{table}[t!]
  \centering
  \small
  \caption{Evaluation metrics used in Track 2: Event-Based Image Retrieval.}
  \vspace{-3mm}
  \label{tab:metrics_task2}
  \begin{tabularx}{\columnwidth}{@{} l Y @{}}
    \toprule
    \textbf{Metric}   & \textbf{Description}                                                \\
    \midrule
    mAP               & Mean Average Precision—measures overall retrieval precision        \\
    MRR               & Mean Reciprocal Rank—measures how early the first correct item is retrieved \\
    Recall@1          & Whether the correct item is ranked first                            \\
    Recall@5          & Whether the correct item is within the top 5 results                \\
    Recall@10         & Whether the correct item is within the top 10 results               \\
    \bottomrule
  \end{tabularx}
\end{table}

\subsection{Evaluation Metrics}

For both tasks, we propose and select carefully a variation of metrics for the overall evaluation and fair performance ranking between each team. To fairly and comprehensively evaluate submissions in both tracks of our challenge, we propose a combined score calculation protocol by assigning a weight tensor $\mathbf{w}$ to the component score and computing the \textbf{weighted harmonic mean} of the score as the final overall score for each valid submission. The formula to compute the overall score is provided as follows:
\begin{equation}
    \label{eq:metrics_overall}
    \texttt{Overall Score} = r \cdot \frac{\sum_{i} w_i}{\Big( \sum_{i} \frac{w_i}{\texttt{score}_i + \epsilon} \Big)},
\end{equation}
where $\epsilon = 10^{-5}$ is a predefined constant to avoid diminishing denominator during the calculation, and $r$ is the number of valid responses in the submission.

\subsubsection{Track 1: Event-Enriched Image Retrieval and Captioning}

For track 1, we employed two separate sets of metrics classified based on their specific domains - Image Retrieval and Image Captioning. The metrics are provided as in Table \ref{tab:metrics_task1}. The overall score of submissions are calculated based on Equation \ref{eq:metrics_overall}. After careful experiments and overall evaluation under different settings, we chose the set of weights
\begin{align*}
    \mathbf{w} &= [w_{\texttt{AP}},\ 
                    w_{\texttt{R@1}},\ 
                    w_{\texttt{R@10}},\ 
                    w_{\texttt{CLIPScore}},\ 
                    w_{\texttt{CIDEr}}] \\
               &= [0.1,\ 
                    0.2,\ 
                    0.2,\ 
                    0.3,\ 
                    0.2]
\end{align*}

\subsubsection{Track 2: Event-Based Image Retrieval}

In track 2, we chose five metrics to compute the retrieval scores in a similar fashion to track 1. The chosen metrics are described in Table \ref{tab:metrics_task2}. After careful analyses, we pre-determined the set of weights for this track to be
\begin{align*}
    \mathbf{w} &= [w_{\texttt{mAP}},\ 
                    w_{\texttt{MRR}},\ 
                    w_{\texttt{R@1}},\ 
                    w_{\texttt{R@5}},\ 
                    w_{\texttt{R@10}}] \\
               &= [0.3,\ 
                    0.2,\ 
                    0.2,\ 
                    0.15,\ 
                    0.15]
\end{align*}

\subsection{Dataset}

\begin{figure}[t!]
    \centering
    \includegraphics[width=\linewidth]{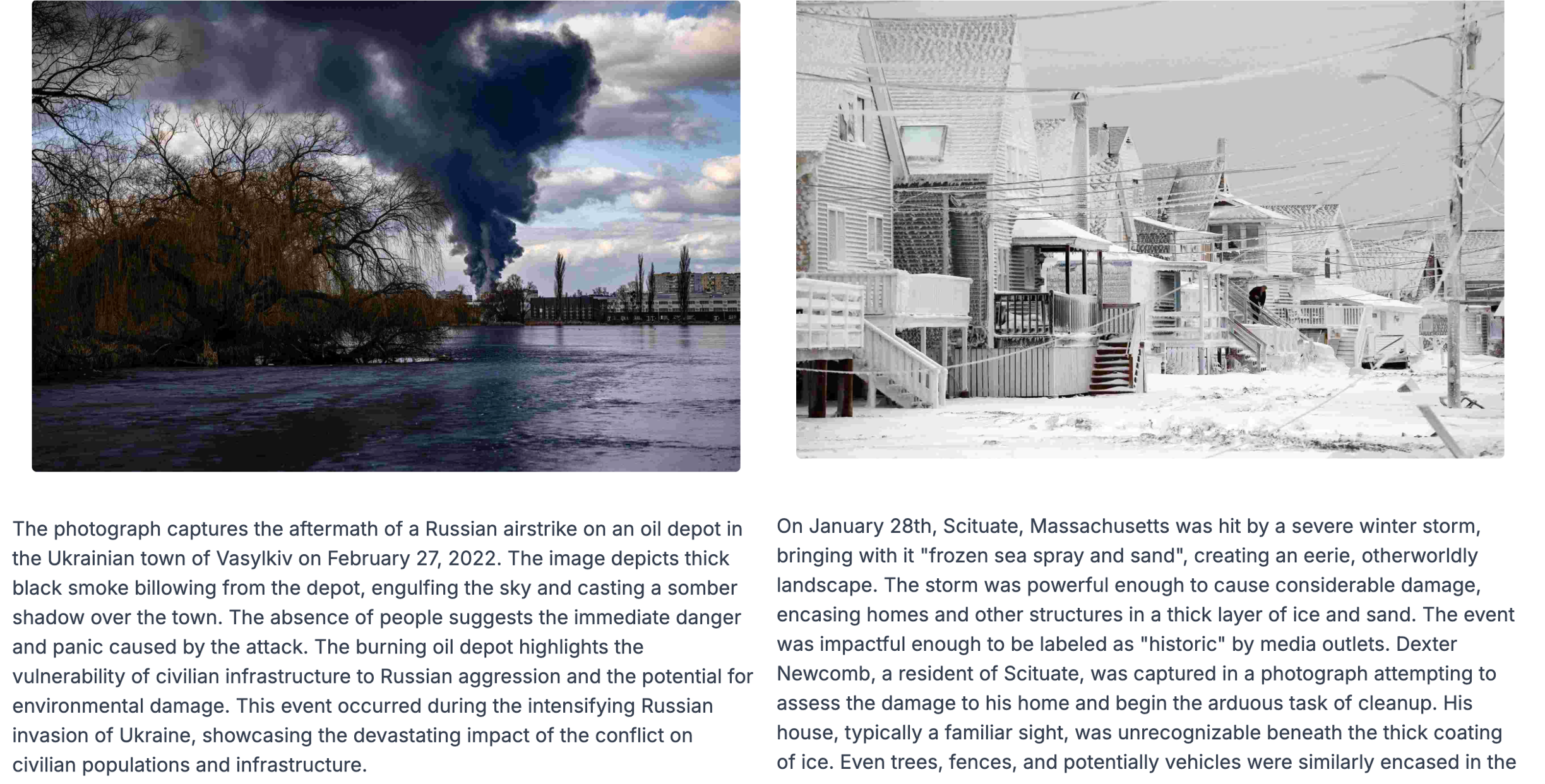}
    \vspace{-7mm}
    \caption{Samples in OpenEvents V1 dataset \cite{openeventsv1}.}
    \label{fig:samples}
\end{figure}

\begin{figure}[t!]
    \centering
    \includegraphics[width=\linewidth]{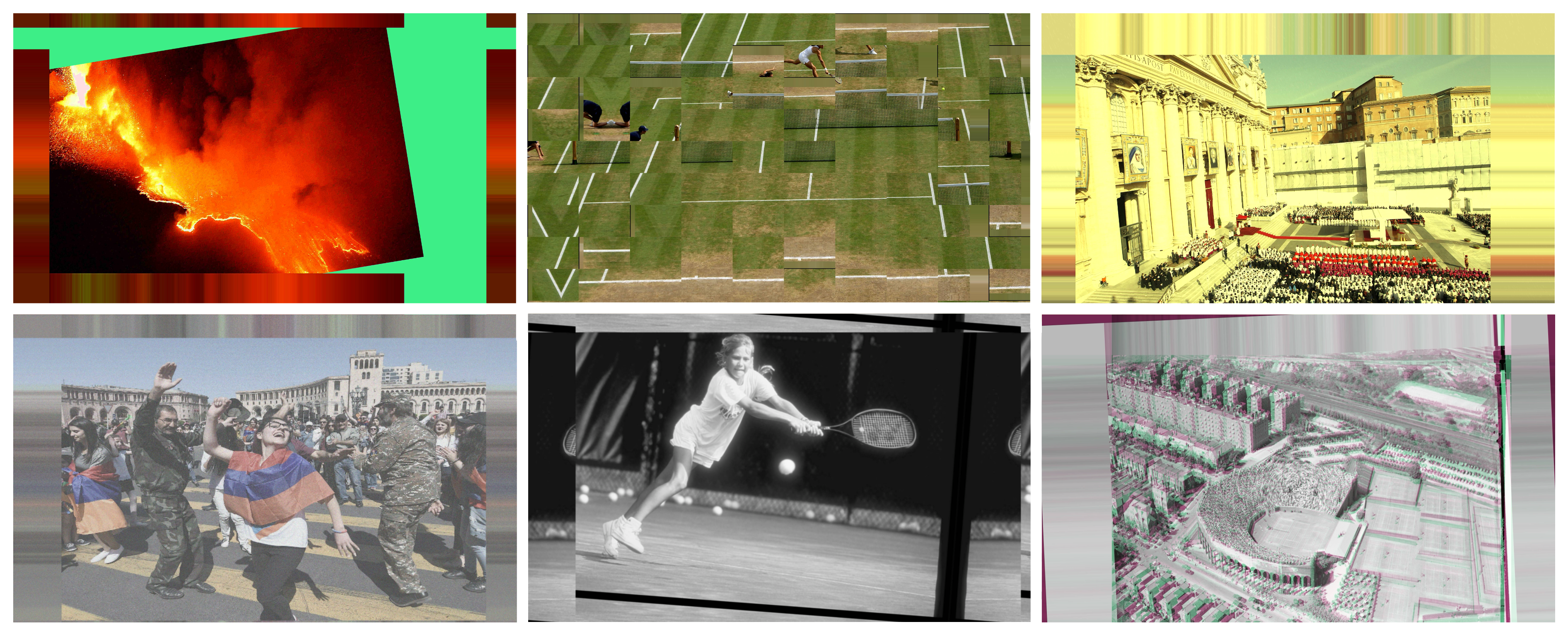}
    \vspace{-5mm}
    \caption{Image query in the Private Set of Track 1.}
    \label{fig:query_private_set}
\end{figure}

The EVENTA 2025 Grand Challenge utilizes the OpenEvents V1 dataset~\cite{openeventsv1}, a large-scale, event-driven corpus designed to bridge the gap between visual content and real-world news understanding. Collected from more than a decade of reporting by two major international outlets, CNN and The Guardian, the dataset captures the dynamic intersection of images, events, and storytelling (\autoref{fig:samples}). It comprises over 200,000 news articles paired with more than 400,000 images, spanning the years 2011–2022 and covering diverse domains such as politics, climate, technology, culture, and sports. To support rigorous evaluation, the dataset also includes more than 30,000 annotated image–event caption pairs, curated and organized into training, public test, and private test splits. This combination of scale, diversity, and event-centric annotation provides a robust foundation for benchmarking models that seek to advance event-enriched image captioning and retrieval. We remark that image queries in the Private Set of Track 1 were heavily augmented to simulate various scenarios in the real-life (\autoref{fig:query_private_set}).

\subsection{Rules}

The EVENTA challenge is conducted under the following rules to ensure fairness, reproducibility, and scientific integrity:

\begin{figure}[t!]
    \centering
    \includegraphics[width=\linewidth]{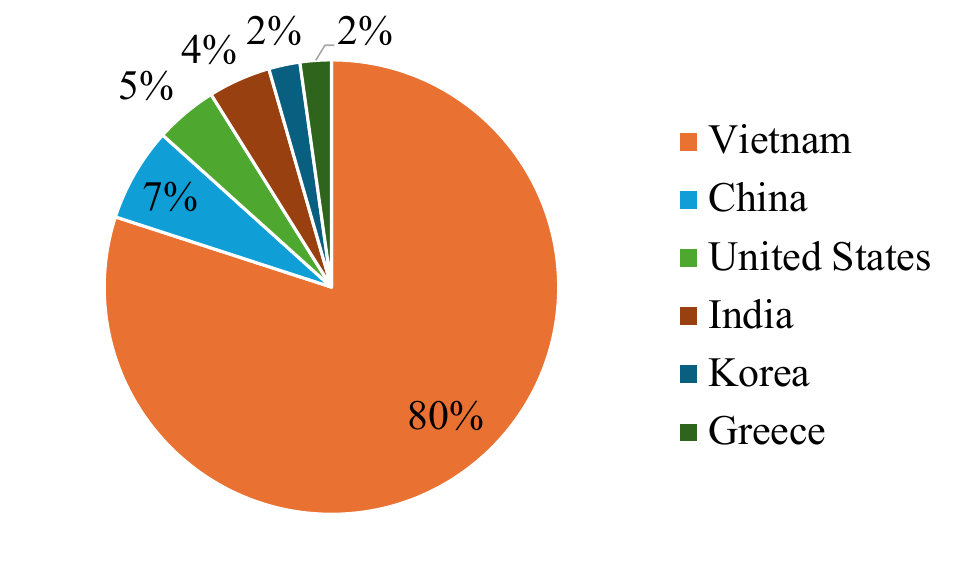}
    \vspace{-10mm}
    \caption{Demographics of participating teams in the EVENTA 2025 Grand Challenge.}
    \label{fig:participants_chart}
\end{figure}

\begin{itemize}
    \item Participants are not allowed to annotate the test sets.
    \item External datasets are permitted. However, participants may only use publicly accessible datasets and pre-trained models. The use of private datasets or pre-trained models is strictly prohibited.
    \item Open sources are allowed. However, commercial tools, libraries, APIs, etc. are strictly prohibited.
    \item Final scores are calculated based on performance on the Private Test set.
    \item Participants must make their source code publicly available on GitHub to ensure reproducibility.
    \item Participants must submit a detailed paper through the official challenge platform before the deadline to validate their solutions.
    \item Late submissions would not be accepted.
    \item Only registered teams that submit papers are eligible to win, but all participants’ scores are recognized.
\end{itemize}

\subsection{Participants}

The challenge attracted 45 registered teams representing six countries. The majority of participants came from Vietnam, with 36 teams from institutions including the University of Science, the University of Information Technology, Ha Noi University of Science and Technology, and Science Lab. Additional participation included one team from Greece (National Technical University of Athens), two teams from the United States (Northwestern University and Hitachi America Research \& Development), three teams from China (Harbin Institute of Technology, Fudan University, and South China University of Technology), one team from Korea (Kookmin University), and two teams from India (Jaypee Institute of Information Technology and Indian Institute of Technology Dhanbad). This diverse participation highlights the strong international interest in advancing event-enriched multimedia understanding.

\section{Challenge Results}

\begin{table}[t!]
\caption{Performance of top teams in the Private Set of Track 1.}
\vspace{-3mm}
\label{tab:track1}
\resizebox{\columnwidth}{!}{
\begin{tabular}{c|l|c|ccc|cc}
\toprule
\textbf{Rank} & \textbf{Team} & \textbf{Overall} & \textbf{AP} & \textbf{R@1} & \textbf{R@10} & \textbf{CLIPScore} & \textbf{CIDEr} \\
\midrule
1 & Cerebro & 0.550 & 0.991 & 0.989 & 0.995 & 0.826 & 0.210 \\
2 & SodaBread & 0.547 & 0.982 & 0.977 & 0.988 & 0.870 & 0.205 \\
3 & Re:zero Slavery & 0.451 & 0.955 & 0.945 & 0.973 & 0.732 & 0.156 \\
4 & ITxTK9 & 0.420 & 0.966 & 0.955 & 0.983 & 0.828 & 0.133 \\
5 & HCMUS-NoName & 0.282 & 0.708 & 0.663 & 0.801 & 0.783 & 0.081 \\
\bottomrule
\end{tabular}
}
\end{table}

\subsection{Track 1: Event-Enriched Image Retrieval and Captioning}

Track 1 evaluated approaches on retrieving relevant contextual evidence and generating event-enriched captions. Most teams converged on a retrieval-augmented captioning paradigm, where external article evidence was first retrieved and re-ranked, then fused with visual features for caption generation. Common design traits included the use of strong vision–language embeddings for candidate retrieval, multi-source contextual signals to enrich captions, and large language models guided by structured prompting. While these approaches achieved near-perfect retrieval accuracy (AP and Recall@1 above 0.95 for the top teams), CIDEr scores remained modest, revealing that fluent and contextually faithful caption generation is still the bottleneck (\autoref{tab:track1}).

Among the top performers, \textbf{Cerebro} secured first place with an overall score of 0.550. Its technical strength lies in uncertainty-guided re-ranking, using entropy, distance, and similarity heuristics, to refine candidate selection, coupled with chain-of-thought prompting that integrates five contextual sources (article summaries, QA pairs, named entities, generated captions, and crawled captions). This comprehensive reasoning pipeline enabled narrative-rich captions and near-perfect retrieval (AP = 0.991, Recall@1 = 0.989). \textbf{SodaBread}, a close second with 0.547, emphasized precision at both retrieval and captioning stages by pairing DINOv2 global similarity with patch-level nearest neighbor matching, followed by multi-source context fusion in Qwen3. Its Semantic Gaussian Normalization module further improved factual grounding and fluency, yielding the highest CLIPScore (0.870) among all teams. \textbf{Re:zero Slavery}, ranked third with 0.451, combined BEiT-3 and SigLIP retrieval with ORB and SIFT re-ranking, and used InstructBLIP plus a fine-tuned Qwen3 (via QLoRA) to generate paragraph-level enriched captions that strongly emphasized temporal cues and named entities.

\subsection{Track 2: Event-Based Image Retrieval}

Track 2 focused on retrieving the most relevant images given realistic event-related captions, a task that requires bridging complex textual event descriptions with visually grounded evidence. All leading teams adopted multi-stage approaches that share several common design principles: (1) an initial article retrieval step using strong pretrained language or multimodal embeddings to narrow the candidate pool, (2) a reranking stage to refine results using additional semantic or contextual signals, and (3) an image-level matching step that integrates vision–language models for cross-modal alignment. Moreover, most approaches employed fusion strategies such as Reciprocal Rank Fusion (RRF) to combine outputs from multiple retrieval signals, highlighting a trend toward robustness through ensemble-style aggregation. While Recall@10 exceeded 0.70 for the top teams, Recall@1 remained relatively low, underscoring the persistent challenge of fine-grained semantic grounding between narrative-rich captions and precise visual matches (\autoref{tab:track2}).

Despite these shared strategies, the top three teams distinguished themselves through their specific techniques. \textbf{NoResources}, the winning system, leveraged a multi-stage framework where Qwen3-embedding handled article retrieval, Qwen3-reranker performed event-aware reranking, and Qwen2-VL scored caption–image similarity, with RRF providing robustness. This design yielded the highest overall score (0.577), demonstrating strong adaptability and balanced performance across all metrics. \textbf{23Trinitrotoluen}, ranked second with a score of 0.572, introduced a hybrid dense–sparse strategy by integrating OpenCLIP, Nomic embeddings, and BM25 retrieval, enhanced with semantic chunking and boosting for re-ranking. This balance of semantic richness and lexical precision proved highly effective in large-scale event retrieval. \textbf{LastSong}, finishing third with 0.563, emphasized hierarchical text modeling by combining sparse, chunked, long-context, and document-level embeddings for article retrieval, followed by reranking and vision-language alignment for image matching. Its fusion of article and image similarity enabled competitive performance, with only a narrow margin separating it from the top two teams.

\begin{table}[t!]
\caption{Performance of top teams in the Private Set of Track 2.}
\vspace{-3mm}
\label{tab:track2}
\resizebox{\columnwidth}{!}{
\begin{tabular}{c|l|c|ccccc}
\toprule
\textbf{Rank} & \textbf{Team} & \textbf{Overall} & \textbf{mAP} & \textbf{MRR} & \textbf{R@1} & \textbf{R@5} & \textbf{R@10} \\
\midrule
1 & NoResources & 0.577 & 0.563 & 0.563 & 0.469 & 0.690 & 0.744 \\
2 & 23Trinitrotoluen & 0.572 & 0.558 & 0.558 & 0.456 & 0.698 & 0.762 \\
3 & LastSong & 0.563 & 0.549 & 0.549 & 0.449 & 0.695 & 0.738 \\
4 & Sharingan Retrievers & 0.533 & 0.521 & 0.521 & 0.428 & 0.640 & 0.705 \\
5 & ZJH-FDU & 0.368 & 0.361 & 0.361 & 0.270 & 0.491 & 0.525 \\
\bottomrule
\end{tabular}
}
\end{table}

Overall, the results across both tracks demonstrate the feasibility of event-enriched captioning and retrieval while also revealing ongoing challenges. Top-performing teams leveraged retrieval-augmented strategies and multimodal fusion to achieve strong performance, but gaps remain in generating fluent, contextually accurate captions and in handling fine-grained event-level retrieval. These findings highlight promising directions for future research, including improved integration of external knowledge sources, robust evaluation metrics, and models capable of reasoning over abstract or ambiguous event queries.

\section{Research Impact}

The EVENTA 2025 Grand Challenge establishes a novel research agenda in multimedia understanding by advancing from surface-level description toward event-enriched interpretation of visual content. Unlike traditional captioning and retrieval tasks that emphasize objects, scenes, or actions in isolation, EVENTA explicitly integrates contextual dimensions such as participants, temporal and spatial information, outcomes, and significance. This event-driven perspective positions the challenge at the intersection of computer vision, natural language processing, and information retrieval, encouraging the development of models that can generate narrative-rich and semantically grounded outputs.

EVENTA Challenge also provides the research community with a platform to explore new approaches for RAG, multimodal fusion, and knowledge-grounded captioning. By prioritizing factual accuracy, contextual depth, and narrative coherence, the challenge also promotes evaluation protocols that move beyond conventional automatic metrics, driving the community toward more comprehensive and human-centered assessments of multimodal understanding.

The broader significance of our EVENTA Challenge extends to multiple real-world applications, including journalism, digital archiving, media search, accessibility, and cultural preservation. By demonstrating the feasibility of systems capable of answering the fundamental who, when, where, what, and why questions of an image, the challenge lays the foundation for next-generation multimedia technologies that are more context-aware, interpretable, and socially relevant. Through strong international participation and interdisciplinary engagement, we foster a vibrant community of researchers and practitioners dedicated to advancing event-driven multimedia AI.

\section{Conclusion and Future Directions}

The EVENTA 2025 Grand Challenge introduced the first large-scale benchmark for event-enriched image captioning and retrieval, addressing the limitations of conventional visual understanding by explicitly emphasizing contextual, narrative, and event-level semantics. With participation from 45 teams across six countries, the challenge demonstrated both the feasibility of new tasks and the strong global interest in advancing event-aware multimodal intelligence. Participant submissions underscored the potential of RAG and multimodal fusion, while also highlighting important challenges that remain related to factual grounding, narrative coherence, and generalization across diverse event domains.

EVENTA opens several promising avenues for future research. Expanding the dataset to incorporate additional modalities such as video, audio, or social media streams would enable richer and more dynamic event understanding. Developing evaluation protocols that better capture factual accuracy, contextual depth, and narrative quality remains an important direction, as current metrics often fail to reflect human judgments of informativeness and coherence. Furthermore, the integration of structured knowledge bases, temporal reasoning, and causal inference has the potential to enhance system performance in capturing the who, when, where, what, and why dimensions of events. Finally, continued collaboration across computer vision, natural language processing, information retrieval, and the digital humanities will be essential to advancing the broader research agenda of context-aware, event-driven multimedia AI.

Through EVENTA 2025 Grand Challenge, we have taken a decisive step toward bridging the gap between vision and storytelling. We hope this challenge will inspire the community to pursue the development of systems that move beyond describing what is visible to interpreting the meaning, implications, and human significance of visual media.

\begin{acks}
This research is supported by research funding from Faculty of Information Technology, University of Science, Vietnam National University - Ho Chi Minh City.


 We would like to thank the ACM Multimedia 2025  organizers for agreeing to host our EVENTA Challenge and for their support. We also thank the Codabench open source competition platform for hosting our leaderboard submissions.

\end{acks}


\bibliographystyle{ACM-Reference-Format}
\balance
\bibliography{ref}


\begin{thebibliography}{18}


\ifx \showCODEN    \undefined \def \showCODEN     #1{\unskip}     \fi
\ifx \showISBNx    \undefined \def \showISBNx     #1{\unskip}     \fi
\ifx \showISBNxiii \undefined \def \showISBNxiii  #1{\unskip}     \fi
\ifx \showISSN     \undefined \def \showISSN      #1{\unskip}     \fi
\ifx \showLCCN     \undefined \def \showLCCN      #1{\unskip}     \fi
\ifx \shownote     \undefined \def \shownote      #1{#1}          \fi
\ifx \showarticletitle \undefined \def \showarticletitle #1{#1}   \fi
\ifx \showURL      \undefined \def \showURL       {\relax}        \fi
\providecommand\bibfield[2]{#2}
\providecommand\bibinfo[2]{#2}
\providecommand\natexlab[1]{#1}
\providecommand\showeprint[2][]{arXiv:#2}

\bibitem[Alayrac et~al\mbox{.}(2022)]%
        {alayrac2022flamingo}
\bibfield{author}{\bibinfo{person}{Jean-Baptiste Alayrac}, \bibinfo{person}{Jeff Donahue}, {et~al\mbox{.}}} \bibinfo{year}{2022}\natexlab{}.
\newblock \showarticletitle{Flamingo: a Visual Language Model for Few-Shot Learning}. In \bibinfo{booktitle}{\emph{Advances in Neural Information Processing Systems}}.
\newblock


\bibitem[Biten et~al\mbox{.}(2019)]%
        {biten2019good}
\bibfield{author}{\bibinfo{person}{Ali~Furkan Biten}, \bibinfo{person}{Lluis Gomez}, \bibinfo{person}{Mar{\c{c}}al Rusinol}, {and} \bibinfo{person}{Dimosthenis Karatzas}.} \bibinfo{year}{2019}\natexlab{}.
\newblock \showarticletitle{Good news, everyone! context driven entity-aware captioning for news images}. In \bibinfo{booktitle}{\emph{Proceedings of the IEEE/CVF conference on computer vision and pattern recognition}}. \bibinfo{pages}{12466--12475}.
\newblock


\bibitem[Girshick et~al\mbox{.}(2014)]%
        {girshick2014rich}
\bibfield{author}{\bibinfo{person}{Ross Girshick}, \bibinfo{person}{Jeff Donahue}, \bibinfo{person}{Trevor Darrell}, {and} \bibinfo{person}{Jitendra Malik}.} \bibinfo{year}{2014}\natexlab{}.
\newblock \showarticletitle{Rich feature hierarchies for accurate object detection and semantic segmentation}. In \bibinfo{booktitle}{\emph{Proceedings of the IEEE conference on computer vision and pattern recognition}}. \bibinfo{pages}{580--587}.
\newblock


\bibitem[Hessel et~al\mbox{.}(2021)]%
        {hessel2021clipscore}
\bibfield{author}{\bibinfo{person}{Jack Hessel}, \bibinfo{person}{Ari Holtzman}, \bibinfo{person}{Maxwell Forbes}, \bibinfo{person}{Ronan~Le Bras}, {and} \bibinfo{person}{Yejin Choi}.} \bibinfo{year}{2021}\natexlab{}.
\newblock \showarticletitle{{CLIPScore:} A Reference-free Evaluation Metric for Image Captioning}. In \bibinfo{booktitle}{\emph{EMNLP}}.
\newblock


\bibitem[Jia et~al\mbox{.}(2021)]%
        {pmlr-v139-jia21b}
\bibfield{author}{\bibinfo{person}{Chao Jia}, \bibinfo{person}{Yinfei Yang}, \bibinfo{person}{Ye Xia}, \bibinfo{person}{Yi-Ting Chen}, \bibinfo{person}{Zarana Parekh}, \bibinfo{person}{Hieu Pham}, \bibinfo{person}{Quoc~V. Le}, \bibinfo{person}{Yunhsuan Sung}, \bibinfo{person}{Zhen Li}, {and} \bibinfo{person}{Tom Duerig}.} \bibinfo{year}{2021}\natexlab{}.
\newblock \showarticletitle{Scaling Up Visual and Vision-Language Representation Learning With Noisy Text Supervision}. In \bibinfo{booktitle}{\emph{Proceedings of the 38th International Conference on Machine Learning}}, Vol.~\bibinfo{volume}{139}.
\newblock


\bibitem[Krause et~al\mbox{.}(2017)]%
        {krause2017hierarchical}
\bibfield{author}{\bibinfo{person}{Jonathan Krause}, \bibinfo{person}{Justin Johnson}, \bibinfo{person}{Ranjay Krishna}, {and} \bibinfo{person}{Li Fei-Fei}.} \bibinfo{year}{2017}\natexlab{}.
\newblock \showarticletitle{A hierarchical approach for generating descriptive image paragraphs}. In \bibinfo{booktitle}{\emph{Proceedings of the IEEE conference on computer vision and pattern recognition}}. \bibinfo{pages}{317--325}.
\newblock


\bibitem[Li et~al\mbox{.}(2022)]%
        {pmlr-v162-li22n}
\bibfield{author}{\bibinfo{person}{Junnan Li}, \bibinfo{person}{Dongxu Li}, \bibinfo{person}{Caiming Xiong}, {and} \bibinfo{person}{Steven Hoi}.} \bibinfo{year}{2022}\natexlab{}.
\newblock \showarticletitle{{BLIP}: Bootstrapping Language-Image Pre-training for Unified Vision-Language Understanding and Generation}. In \bibinfo{booktitle}{\emph{Proceedings of the 39th International Conference on Machine Learning}} \emph{(\bibinfo{series}{Proceedings of Machine Learning Research}, Vol.~\bibinfo{volume}{162})}. \bibinfo{publisher}{PMLR}, \bibinfo{pages}{12888--12900}.
\newblock


\bibitem[Li et~al\mbox{.}(2021)]%
        {li2021albef}
\bibfield{author}{\bibinfo{person}{Junnan Li}, \bibinfo{person}{Ramprasaath~R. Selvaraju}, \bibinfo{person}{Akhilesh~Deepak Gotmare}, \bibinfo{person}{Shafiq Joty}, \bibinfo{person}{Caiming Xiong}, {and} \bibinfo{person}{Steven C.~H. Hoi}.} \bibinfo{year}{2021}\natexlab{}.
\newblock \showarticletitle{Align Before Fuse: Vision and Language Representation Learning with Momentum Distillation}. In \bibinfo{booktitle}{\emph{Advances in Neural Information Processing Systems}}.
\newblock


\bibitem[Li et~al\mbox{.}(2024)]%
        {li2024evcap}
\bibfield{author}{\bibinfo{person}{Jiaxuan Li}, \bibinfo{person}{Duc~Minh Vo}, \bibinfo{person}{Akihiro Sugimoto}, {and} \bibinfo{person}{Hideki Nakayama}.} \bibinfo{year}{2024}\natexlab{}.
\newblock \showarticletitle{{EVCap}: Retrieval-Augmented Image Captioning with External Visual--Name Memory for Open-World Comprehension}. In \bibinfo{booktitle}{\emph{Proceedings of the IEEE/CVF Conference on Computer Vision and Pattern Recognition (CVPR)}}.
\newblock


\bibitem[Long et~al\mbox{.}(2015)]%
        {long2015fully}
\bibfield{author}{\bibinfo{person}{Jonathan Long}, \bibinfo{person}{Evan Shelhamer}, {and} \bibinfo{person}{Trevor Darrell}.} \bibinfo{year}{2015}\natexlab{}.
\newblock \showarticletitle{Fully convolutional networks for semantic segmentation}. In \bibinfo{booktitle}{\emph{Proceedings of the IEEE conference on computer vision and pattern recognition}}. \bibinfo{pages}{3431--3440}.
\newblock


\bibitem[Nguyen et~al\mbox{.}(2025)]%
        {openeventsv1}
\bibfield{author}{\bibinfo{person}{Hieu Nguyen}, \bibinfo{person}{Phuc-Tan Nguyen}, \bibinfo{person}{Thien-Phuc Tran}, \bibinfo{person}{Minh-Quang Nguyen}, \bibinfo{person}{Tam~V. Nguyen}, \bibinfo{person}{Minh-Triet Tran}, {and} \bibinfo{person}{Trung-Nghia Le}.} \bibinfo{year}{2025}\natexlab{}.
\newblock \showarticletitle{OpenEvents V1: Large-Scale Benchmark Dataset for Multimodal Event Grounding}. In \bibinfo{booktitle}{\emph{ACM International Conference on Multimedia}}.
\newblock


\bibitem[Qu et~al\mbox{.}(2024)]%
        {qu2024news}
\bibfield{author}{\bibinfo{person}{Tingyu Qu}, \bibinfo{person}{Tinne Tuytelaars}, {and} \bibinfo{person}{Marie-Francine Moens}.} \bibinfo{year}{2024}\natexlab{}.
\newblock \showarticletitle{Visually-Aware Context Modeling for News Image Captioning}. In \bibinfo{booktitle}{\emph{Proceedings of the 2024 Conference of the North American Chapter of the Association for Computational Linguistics}}.
\newblock


\bibitem[Radford et~al\mbox{.}(2021)]%
        {pmlr-v139-radford21a}
\bibfield{author}{\bibinfo{person}{Alec Radford}, \bibinfo{person}{Jong~Wook Kim}, \bibinfo{person}{Chris Hallacy}, \bibinfo{person}{Aditya Ramesh}, \bibinfo{person}{Gabriel Goh}, \bibinfo{person}{Sandhini Agarwal}, \bibinfo{person}{Girish Sastry}, \bibinfo{person}{Amanda Askell}, \bibinfo{person}{Pamela Mishkin}, \bibinfo{person}{Jack Clark}, \bibinfo{person}{Gretchen Krueger}, {and} \bibinfo{person}{Ilya Sutskever}.} \bibinfo{year}{2021}\natexlab{}.
\newblock \showarticletitle{Learning Transferable Visual Models From Natural Language Supervision}. In \bibinfo{booktitle}{\emph{Proceedings of the 38th International Conference on Machine Learning}}, Vol.~\bibinfo{volume}{139}. \bibinfo{publisher}{PMLR}, \bibinfo{pages}{8748--8763}.
\newblock


\bibitem[Ramos et~al\mbox{.}(2023)]%
        {ramos2023extra}
\bibfield{author}{\bibinfo{person}{Rita Ramos}, \bibinfo{person}{Desmond Elliott}, {and} \bibinfo{person}{Bruno Martins}.} \bibinfo{year}{2023}\natexlab{}.
\newblock \showarticletitle{Retrieval-Augmented Image Captioning}. In \bibinfo{booktitle}{\emph{Proceedings of the 17th Conference of the European Chapter of the Association for Computational Linguistics}}. \bibinfo{pages}{3648--3663}.
\newblock


\bibitem[Vedantam et~al\mbox{.}(2015)]%
        {Vedantam_2015_CVPR}
\bibfield{author}{\bibinfo{person}{Ramakrishna Vedantam}, \bibinfo{person}{C. Lawrence~Zitnick}, {and} \bibinfo{person}{Devi Parikh}.} \bibinfo{year}{2015}\natexlab{}.
\newblock \showarticletitle{CIDEr: Consensus-Based Image Description Evaluation}. In \bibinfo{booktitle}{\emph{Proceedings of the IEEE Conference on Computer Vision and Pattern Recognition (CVPR)}}.
\newblock


\bibitem[Wang et~al\mbox{.}(2022)]%
        {wang2022git}
\bibfield{author}{\bibinfo{person}{Jianfeng Wang}, \bibinfo{person}{Zhengyuan Yang}, \bibinfo{person}{Xiaowei Hu}, \bibinfo{person}{Linjie Li}, \bibinfo{person}{Kevin Lin}, \bibinfo{person}{Zhe Gan}, \bibinfo{person}{Zicheng Liu}, \bibinfo{person}{Ce Liu}, {and} \bibinfo{person}{Lijuan Wang}.} \bibinfo{year}{2022}\natexlab{}.
\newblock \showarticletitle{{GIT}: A Generative Image-to-text Transformer for Vision and Language}.
\newblock \bibinfo{journal}{\emph{arXiv:2205.14100}} (\bibinfo{year}{2022}).
\newblock


\bibitem[Xu et~al\mbox{.}(2015)]%
        {xu2015show}
\bibfield{author}{\bibinfo{person}{Kelvin Xu}, \bibinfo{person}{Jimmy Ba}, \bibinfo{person}{Ryan Kiros}, \bibinfo{person}{Kyunghyun Cho}, \bibinfo{person}{Aaron Courville}, \bibinfo{person}{Ruslan Salakhudinov}, \bibinfo{person}{Rich Zemel}, {and} \bibinfo{person}{Yoshua Bengio}.} \bibinfo{year}{2015}\natexlab{}.
\newblock \showarticletitle{Show, attend and tell: Neural image caption generation with visual attention}. In \bibinfo{booktitle}{\emph{International conference on machine learning}}. PMLR, \bibinfo{pages}{2048--2057}.
\newblock


\bibitem[Yang et~al\mbox{.}(2023)]%
        {yang2023alip}
\bibfield{author}{\bibinfo{person}{Kaicheng Yang}, \bibinfo{person}{Jiankang Deng}, \bibinfo{person}{Xiang An}, \bibinfo{person}{Jiawei Li}, \bibinfo{person}{Ziyong Feng}, \bibinfo{person}{Jia Guo}, \bibinfo{person}{Jing Yang}, {and} \bibinfo{person}{Tongliang Liu}.} \bibinfo{year}{2023}\natexlab{}.
\newblock \showarticletitle{{ALIP}: Adaptive Language-Image Pre-Training with Synthetic Caption}. In \bibinfo{booktitle}{\emph{Proceedings of the IEEE/CVF International Conference on Computer Vision (ICCV)}}.
\newblock


\end{thebibliography}

\end{document}